\documentclass[runningheads]{llncs}
\usepackage{amsmath}
\usepackage{amssymb}
\usepackage{caption}
\usepackage{subcaption}
\usepackage{multirow}
\usepackage[T1]{fontenc}
\usepackage{graphicx}
\usepackage{booktabs}
\usepackage{algorithm}
\usepackage{algpseudocode}
\usepackage{tikz}
\usepackage{hyperref}
\hypersetup{
    colorlinks=true,
    urlcolor=blue,
    }
\begin{document}
\title{Image Distillation for Safe Data Sharing in Histopathology}
%
%\titlerunning{Abbreviated paper title}
% If the paper title is too long for the running head, you can set
% an abbreviated paper title here
%
\author{Zhe Li\orcidID{0009-0003-3101-718X} \and
Bernhard Kainz\orcidID{0000-0002-7813-5023}}
\authorrunning{}
% First names are abbreviated in the running head.
% If there are more than two authors, 'et al.' is used.
%
\institute{Image Data Exploration and Analysis Lab, Friedrich-Alexander-Universität Erlangen-Nürnberg \\
\email{zhe.li@fau.de}
}

\maketitle              % typeset the header of the contribution

\begin{abstract}

Histopathology can help clinicians make accurate diagnoses, determine disease prognosis, and plan appropriate treatment strategies. As deep learning techniques have developed successfully in the medical domain, the primary challenges become limited data availability and concerns about data sharing and privacy. Federated learning has addressed this challenge by training models locally and updating parameters on a server. However, issues, such as domain shift and bias, persist and impact overall performance. Dataset distillation presents an alternative approach to overcoming these challenges. It involves creating a small synthetic dataset that encapsulates essential information, which can be shared without constraints. At present, this paradigm is not practicable as current distillation approaches only generate non human readable  representations and exhibit insufficient performance for downstream learning tasks.
We train a latent diffusion model and construct a new distilled synthetic dataset with a small number of human readable synthetic images. Selection of maximally informative synthetic images is done via graph community analysis of the representation space. We compare downstream classification models trained on our synthetic images to models trained on real images and achieve performances suitable for practical application. Codes are available at \url{https://github.com/ZheLi2020/InfoDist}.

\end{abstract}

\section{Introduction}
In histopathology, pathologists examine thin sections of tissues derived from biopsies. These tissue samples are typically stained to enhance the visibility of cellular structures. Subsequently, the stained slides undergo microscopic analysis to identify abnormal changes or patterns that may indicate the presence of a disease.
Histopathology plays a crucial role in understanding various medical conditions, help clinicians make accurate diagnoses, determine disease prognosis, and plan appropriate treatment strategies.

Deep learning and computer vision methods have demonstrated success in the analysis of histopathology images, encompassing tasks such as disease detection, tumor classification, and cell segmentation. Some approaches use multiple instance learning (MIL)~\cite{Yang22-remixMIL} or hierarchical approaches~\cite{chen2022scaling,guan2022node,jiang2023hierarchical}. These achieve commendable classification performance. Nevertheless, the challenges of data scarcity and data sharing is hard to tackle due to concerns related to privacy protection, diverse dataset formats, and legal and regulatory frameworks. Federated learning has addressed this issue through a distributed approach, where individuals train their data locally, and model parameters are updated on a central server. While this method maintains data privacy, it introduces new challenges such as domain shift and bias.

Some researchers have explored knowledge distillation~\cite{azadi2023all,Yu23-prototypedistillation,wang2023black,yu2023slpd,zhong2023semi} or the generation of synthetic images~\cite{shrivastava2023nasdm,ye2023synthetic,kang2023one} as data enrichment methods to address the data shortage problem and enhance performance.  Concerns about data privacy still exist because these methods only assist the training and models are trained on real data as before. In this context, dataset distillation emerges as a more appropriate solution. It learns from a large real dataset and generating a small synthetic dataset that maximally encapsulates essential information. Subsequently, downstream tasks are trained on the dataset that solely consists of synthetic images.
Effective dataset distillation would allow sharing of a small synthetic dataset that can a) guarantee to not contain any privacy concerning identifiable information, b) be used as a direct representation of the underlying data distribution characteristics for downstream applications and bias mitigation, and c) allow resource efficient training and refinement of, \emph{e.g.}, general foundation models. 
Leveraging data distillation to work with compact, synthesized images across clinical sites could effectively anonymize the training process by removing all patient-specific information on the image level.

We would expect that approaches for dataset distillation can achieve comparable performance to those trained on a large real dataset.  
However, the performance achieved by training solely on a small synthetic dataset is sub-optimal.
One reason is the majority of approaches only generate a limited number of synthetic images, typically around $1$ or $10$ for each class. Hence, its practical applicability in the real world is currently limited. 
Our goal is to extract maximum information from the original large dataset with real images, and generate a minimal synthetic dataset that can achieve comparable performance when training a task-specific model. To achieve this, we train a class conditional latent diffusion model and generate $1000$ synthetic images for each class. We then utilize the Infomap algorithm~\cite{blocker2023map} to select $100$ images from it in the representation space of convolutional networks. Afterwards, we train classifiers on this distilled subset to evaluate the set's representational power.
This approach enables us to achieve comparable classification accuracy to real images while efficiently reducing storage costs and training efforts simultaneously.

To the best of our knowledge, there are no works that have specifically investigated dataset distillation for histopathology. Our contribution consists of:
\begin{enumerate}
  \item We propose an approach to use synthetic images for data sharing, which mitigates privacy concerns -- InfoDist. 
  \item We explore an efficient method to select images with essential information. The process involves projecting images to the embedding using a pre-trained convolutional network, followed by the utilization of the map equation and Infomap algorithm~\cite{blocker2023map} to select images with high modular centrality.
  \item We design a contrasting learning loss to further improve performance.
  \item In our experiments, we report the average results of test accuracy, F1 score, and AUC score. We demonstrate that competitive performance can be achieved when training solely on a distilled set of synthetic images.
\end{enumerate}

\noindent\textbf{Related Work.} 
Dataset distillation is initially explored on natural images, such as CIFAR-10 or ImageNet to reduce the storage and computation cost. Recently, such frameworks have been applied on gastric X-ray images~\cite{li2022compressed} and a COVID-19 chest X-ray dataset~\cite{li2022dataset}. There is no work on histopathology images so far. Most researchers focus on approaches~\cite{tang2023multiple,shao2023lnpl,huang2023conslide,qu2023boosting,wang2023iteratively,jin2023gene} using the MIL method, multi-modal data combination~\cite{lu2023multi} or data augmentation~\cite{gadermayr2023mixup} for the classification of whole slide images or segmentation~\cite{deng2023democratizing}.

\section{Method}

\begin{figure}[tb]
   \centering
      \includegraphics[width=1.0\linewidth]{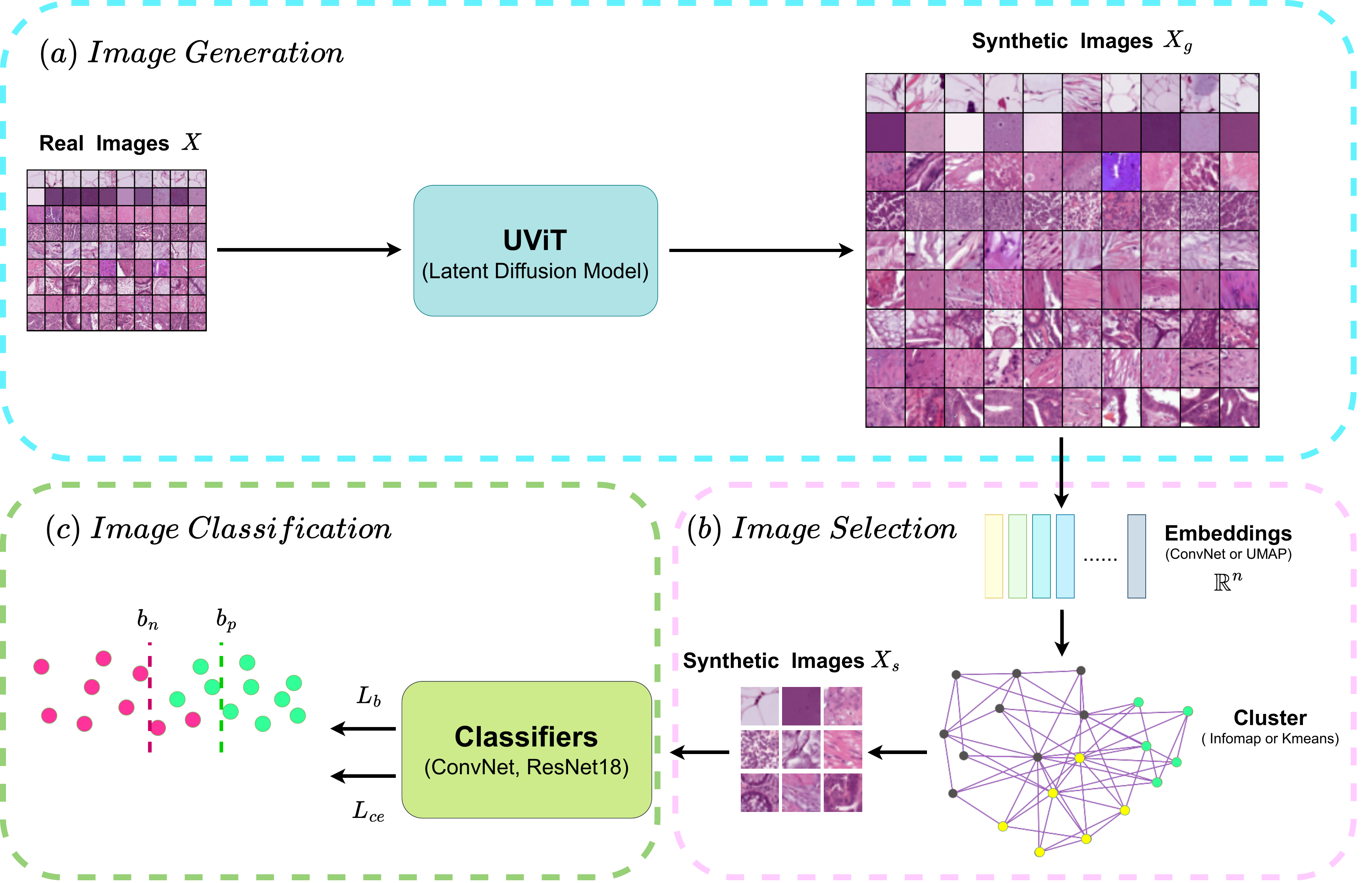}
   \caption{
   Overview of our InfoDist approach. (a) We train a latent diffusion model UViT~\cite{bao2023all} and generate a synthetic dataset. (b) We extract the image embeddings by a pre-trained convolutional network or UMAP~\cite{mcinnes2018umap}, then use the modified infomap algorithm to detect communities. We select a small synthetic dataset in which images have high modular centrality in each community. (c) We train the classifiers only on the small selected synthetic dataset and apply both cross entropy loss $\mathcal{L}_{ce}$ and contrastive learning loss $\mathcal{L}_{con}$ in training.}
   \label{fig:overviewframework}
   \vspace{-8mm}
\end{figure}

Our idea for InfoDist is outlined in Fig.~\ref{fig:overviewframework}. The overall goal is to select the most informative subset of images $X_s$ from a set of all images $X$, so that $X_s$ will provide enough information to train another downstream classifier, ideally yielding the same classification performance on an unseen test set $X_t$ as if trained on the full $X$. $X$ could not readily be shared with third parties, since this may be samples from a set of real patient data. Thus we propose to transform $X$ to a newly generated synthetic set $X_g = \mathcal{F}_{\theta}(X)$ with infinite extent and start the data distillation process from there. 

Latent diffusion models recently became a viable option to model $\mathcal{F}_{\theta}(X)$ without too much loss of information about the underlying distribution in $X$. 
We utilize a class conditional latent diffusion model U-ViT~\cite{bao2023all} to model $X_g = \mathcal{F}_{\theta}(X)$. It's a backbone latent diffusion model in a U-Net shape, where each block is a vision transformer~\cite{dosovitskiy2020image}. 
The diffusion model is trained on real data and learns the data distribution. It can be trained locally once without sharing it publicly. Then we can generate synthetic images, but these images are different from any real images, even though they contain some realistic features. Therefore, there is no personal information in synthetic images.
After training on the pathological dataset, it can generate synthetic images $X_g \in \mathbb{R}^{3\times W \times H}$ that are comparable to real images.
Our goal is to capture the input data distribution as comprehensively as possible. To achieve this, we generate a large number of synthetic images for $X_g$ and then select a small subset of representative images.

$X_g$ can be further embedded into the embedding space $\mathbb{R}^n$ of a classifier if the size of elements $\in X_g$ is computationally prohibitive.
The embedding space $\mathbb{R}^n$ is the output of the penultimate layer of a classifier pretrained on all real images $X$. 

From the embedding space $X_g \in \mathbb{R}^n$ we construct a weighted graph from the set $X_g = {x_1,...,x_N}$ with a metric $d: X \times X \rightarrow \mathbb{R}_{w\geq \eta} $, where $w$ is the inversed Euclidean distance and $\eta$ is a threshold.
This leads to a weighted directed graph $\bar{G} = (V,E,w)$, where the nodes $V = X_g$ and the edges $E = {(x_i, x_{i_{j}}) | w \geq \eta}$.

An alternative embedding method is UMAP~\cite{mcinnes2018umap}. We flatten $X_g$ and reduce the dimension by UMAP which also constructs a graph by Euclidean distance. In this graph, each node $x_i$ connects its top-k neighbors and the edges become $E = {(x_i, x_{i_{j}}) | 1 \leq j \leq k, 1 \leq i \leq N}$. We also use this approach in our experiments for comparison.

We hypothesize that $\bar{G}$ contains essential information about the relevance among individual images $I \in X_g$ for dataset distillation and their effectiveness to shape decision boundaries in downstream classifiers. In exploring the nuances of nodes and the identification of weak boundaries among clusters, we introduce an unsupervised method to unearth samples that encode intricate details, even within a single class. Here, community detection emerges as an important factor; 
specifically, the nodes with the high modular centrality in each community could be deemed the most relevant. 
The scalar score of modular centrality combines two scores to quantify both the intra-community and inter-community influence~\cite{ghalmane2019centrality}.
This allows community detection not merely as a tool for group identification but as a method of compression and dataset distillation. Via the Infomap algorithm~\cite{blocker2023map}, we can identify those key samples that, despite their limited number, encapsulate the dataset's complexity and facilitate the formation of robust decision boundaries, thereby enhancing the efficiency and effectiveness of the distillation process.

Community detection can leverage the Map Equation~\cite{rosvall2009map}, an unsupervised method grounded in information theory, which aims to optimize community identification based on the principle of minimum description length~\cite{grunwald2005advances}. This approach seeks to encapsulate the behavior of a random walker within a network in the most concise way, by reducing the expected per-step codelength. It achieves this by organizing the network into clusters within which the random walker is likely to remain for extended periods. The Map Equation method does not require simulating random walks to achieve its objectives; rather, it analytically computes the codelength:
\begin{equation}
L(M) = q_{\curvearrowright}H(Q) + \sum_{i=1}^{m}p_{\circlearrowright}^iH(P^i)
\end{equation}
where $L(M)$ is the total codelength for a given partition $M$ of the network into $m$ clusters, $q_{\curvearrowright}$ is the probability of the random walker transitioning between clusters, $H(Q)$ is the entropy of the module exit probabilities, quantifying the uncertainty in module exits, $p_{\circlearrowright}^i$ is the probability of the random walker staying within module $i$, $H(P^i)$ is the entropy of the visitation probabilities within module $i$, and $m$ is the number of clusters.
Algorithm~\ref{algo:infomap}, which is a modified version of InfoMap~\cite{blocker2023map} can then be used to distill nodes that have high modular centrality from each community uniformly.
Here, we generate graphs for each class separately, so all synthetic images with same class label are nodes in the graph. We calculate the inverse Euclidean distance and apply $softmax$ operation as weights of links between nodes. To ensure that higher weight represents a stronger relation between nodes, we set a threshold $\eta$ to remove the links with low weights.

\begin{algorithm}
\caption{Our proposed InfoDist algorithm for Community Detection 
}\label{algo:infomap}

\begin{algorithmic}[1]
\State \textbf{Input}: The real images $X^k$ or synthetic images $X_g^k$ with class label $k$; \\
\hskip3.3em $N$, the number of selected images.
%\ \ \ \ \ \ \ \ \ \ 
\State Initialize each node as its own community $C$.
\State Initialize empty distillation set $X_s$. 
\Repeat
    \For{each node $i$ in the network}
        \For{each community $C$ that $i$'s neighbors belong to}
            \State Calculate the change in $L(M)$ if $i$ is moved to $C$.
        \EndFor
        \State Move node $i$ to the community that results in the greatest decrease in $L(M)$.
    \EndFor
    \State Update the community structure based on node movements.
\Until{no further reduction in $L(M)$ is possible}
\State Calculate $N_i$, the number of images that needs to be selected from each community.
\For{each community $C_i$ in the network}
\State{$X_s \leftarrow$ $I \in X^k$ or $I \in X_g^k$ nodes that have high modular centrality.}
\EndFor
\State  Apply the algorithm recursively to images in other classes to construct the small real or synthetic dataset. 
\end{algorithmic}
\end{algorithm}

\noindent\textbf{Contrastive Learning Loss.}
\label{sec:contrastive}
To further the performance of downstream classifiers that are trained on very limited data like distilled datasets, we propose the use of a contrastive learning loss~\cite{yao2023explicit}. We set two boundaries and process output probabilities of the convolutional model during training to generate the boundaries with the assistance of masks defined by the ground truth. 
Specifically, we slice the output to obtain the probabilities for one class and mask out the positive and negative samples within the current batch. 
The mask is generated based on the ground truth label. If the ground truth corresponds to the current processing class label, the value is set to $1$. Otherwise, it is set to~$0$.
After applying the mask to probabilities, we obtain the probabilities of samples associated with the current class label. We assume that there are positive samples with low probabilities which are easily confused with negative samples. We set a threshold such that a percentage $\rho$ of the positive samples are predicted correctly. Thus, we sort the probabilities of positive samples in descending order and set the sample probability at the last position of $B\times \rho$ samples as the positive boundary $b_p$. The remaining $B\times(1-\rho)$ positive samples with lower probabilities are contributed to the loss calculation as the first term of Eq.~\ref{lossbounsingle}.
The negative boundary $b_n$ for those probabilities with a mask value $0$ is calculated by subtracting a hyperparameter $\tau$ from $b_p$. 
The probabilities that are higher than the negative boundary are calculated in the loss as the second term of Eq.~\ref{lossbounsingle}.

\begin{equation}\label{lossbounsingle}
    \mathcal{L}_{c} = \sum_{i=0}^{Pos}|min((p_i - b_p), 0)| + \sum_{j=0}^{Neg}|max((p_j - b_n + \tau), 0)|, 
\end{equation}

where \textit{Pos} denotes the number of positive images for current class label in the ground truth and \textit{Neg} denotes the number of the remaining images with other class labels. $p_i, p_j$ is the output probability of the synthetic images $s_i, s_j$ after $softmax$ operation.
We compute two boundaries and the contrastive learning loss for each class separately and calculate the contrastive loss 
$
    \mathcal{L}_{b} = \sum_{c=0}^{C}\mathcal{L}_{c}
$
Finally, the total loss is a combination with  cross entropy
$\label{losssyn}
    \mathcal{L} = \mathcal{L}_{b} + \mathcal{L}_{ce}.
$

\section{Experiments}

\textbf{Datasets.} We evaluate our distillation approach on the public MedMNIST datasets~\cite{yang2023medmnist} which comprises MNIST-like datasets featuring standardized biomedical images that consists of a total of $12$ datasets for 2D and $6$ datasets for 3D. Our method is specifically applied to one of these datasets, PathMNIST, which includes histopathology images for colon biopsies.  
This dataset consists of $107,180$ image samples which are divided into training ($89,996$), validation ($10,004$), and test ($7,180$) sets. All images are resampled into 28 x 28 (2D) resolution.
The PathMNIST dataset has $9$ types of tissues: ADI, adipose tissue; BACK, background; CRC, colorectal cancer; DEB, debris; HE, hematoxylin–eosin; LYM, lymphocytes; MUC, mucus; MUS, smooth muscle; NCT, National Center for Tumor Diseases; NORM, normal colon mucosa; STR, cancer-associated stroma; TUM, colorectal adenocarcinoma epithelium~\cite{kather2019predicting}.

\noindent\textbf{Metrics.} We report $3$ evaluation metrics, the classification accuracy, the F1 score, and the AUC on the test set of PathMNIST.

\noindent\textbf{Implementation Details.} In the embedding process, images are projected into features with dimensions of either $2048$ or $8192$, corresponding to resolutions of $64\times 64$ or $256\times 256$, using a pre-trained ConvNet or ResNet18. The UMAP projects the images into features with dimension $10$, with a specified number of neighbors set to $10$ and a minimum distance parameter of $0.05$. In Infomap, we can either utilize the graph generated by UMAP or calculate the distance matrix using the Euclidean distance where the weight threshold $\eta$ is set to $0.001$ or $0.004$.
For computation cost, training ConvNet model for $5$ times needs about $20$ minutes and ResNet18 needs about $40$ minutes on a A100 GPU. 

\begin{figure}[t]
\begin{center}
    \begin{subfigure}[b]{0.32\linewidth}
         \centering
         \includegraphics[width=\linewidth]{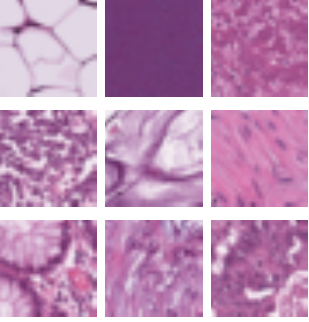}
         \caption{real images}
         \label{}
     \end{subfigure}
     \hfill
     \begin{subfigure}[b]{0.32\linewidth}
         \centering
         \includegraphics[width=\linewidth]{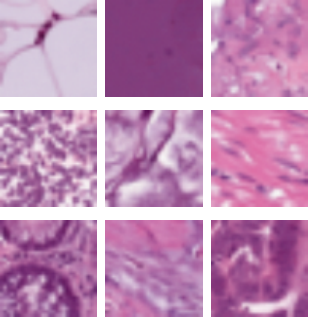}
         \caption{synthetic image $64$}
         \label{}
     \end{subfigure}
     \hfill
     \begin{subfigure}[b]{0.32\linewidth}
        \centering
        \includegraphics[width=\linewidth]{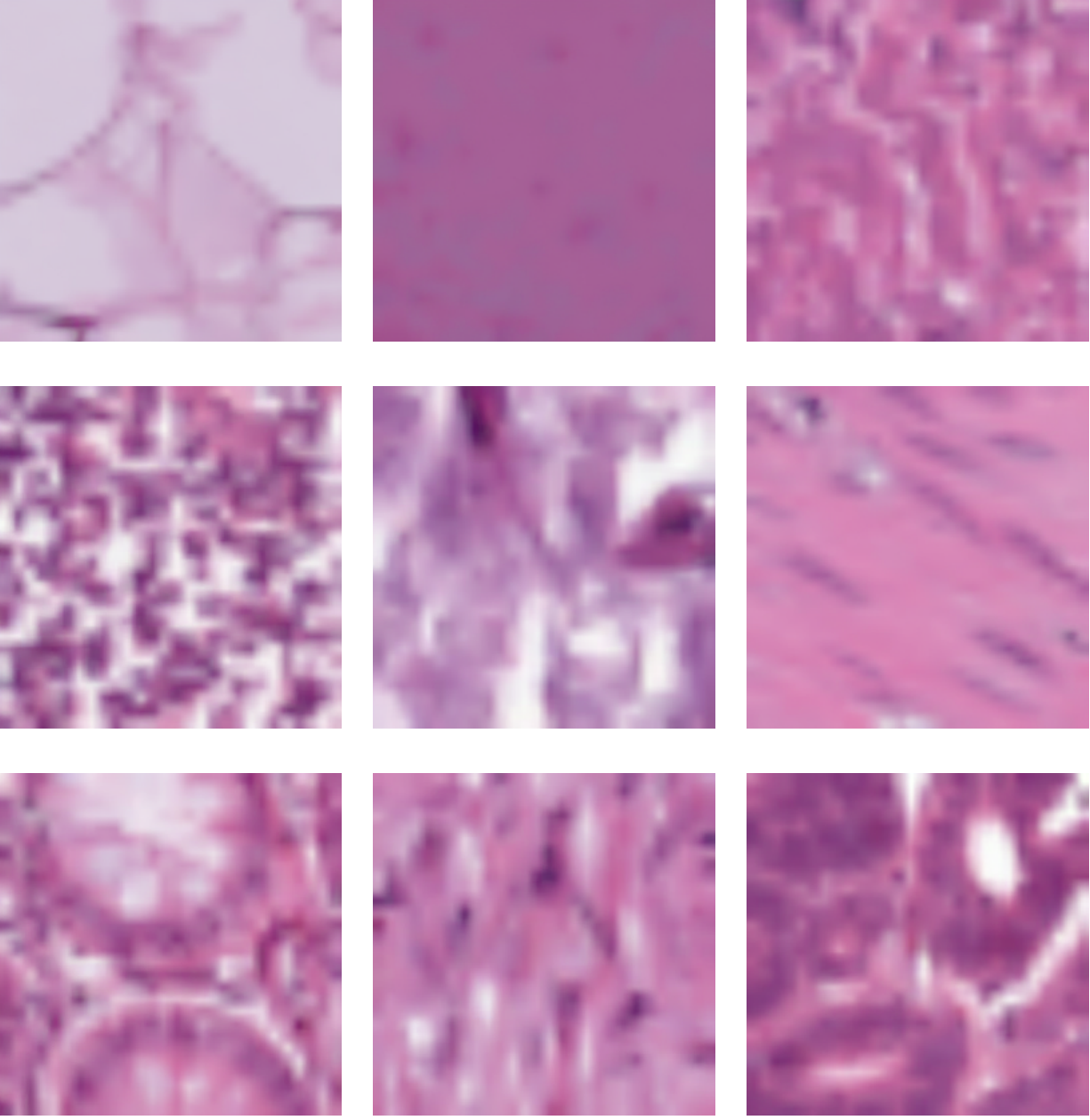}
        \caption{synthetic image $256$}
        \label{}
     \end{subfigure}
\end{center}
\vspace{-4mm}
   \caption{The real and synthetic samples at different resolutions.}
   \vspace{-4mm}
\label{fig:medsamples}
\end{figure}

\noindent\textbf{Synthetic Images.}
We train a class conditional latent diffusion model U-ViT~\cite{bao2023all} on the training set of the PathMNIST dataset which consists of $89,996$ images. We resize input images from resolutions $28\times 28$ to $64\times 64$ or $256\times 256$ respectively before training. After training the U-ViT-L/4 on images with resolution $64\times 64$ and U-ViT-L/2 on $256\times 256$ respectively, we generate $1000$ synthetic images for each class, resulting in a total of $9000$ synthetic images for each size.
Fig.~\ref{fig:medsamples}(a) shows the samples of real images. Fig.~\ref{fig:medsamples}(b) exhibits samples of synthetic images generated at a resolution of $64\times 64$, while Fig.~\ref{fig:medsamples}(c) showcases samples of synthetic images generated at a resolution of $256\times 256$.

\noindent\textbf{Distilled data. }
To create a distilled dataset, we select $100$ images for each class from the generated synthetic dataset with our InfoDist approach. Therefore, the condensed synthetic dataset comprises $900$ images for training a downstream classifier, considering there are $9$ classes in the PathMNIST dataset. 
We employ two classifiers: ConvNet and ResNet18. The reported results are the average of $5$ runs on the entire real test set of the PathMNIST dataset and the corresponding standard deviation. In each run, the selected training images are updated.

Table~\ref{tab:resultcomparesota} shows our results of InfoDist on a small distilled dataset compared to the state-of-the-art. 
We train two classifiers on the whole real training set and indicate results as \textit{Reproduced}. 
We also apply InfoDist on the real dataset and select a small dataset with $900$ real images. We report the results in \textit{Real/real distilled/\-InfoDist}. 
In the bottom part, \textit{Synthetic}, we train two classifiers on all synthetic images and report results in the rows \textit{Reproduced}. For the distilled synthetic dataset with $900$ images, we compare our approach with two baselines, GLAD~\cite{cazenavette2023generalizing} and random image selection.
With resolution $64\times 64$, a ConvNet achieves $69.79$ test accuracy which is comparable to the performance $77.34$ of the distilled real dataset. A ResNet18 can achieve better performance $77.48$ in test accuracy which is on par to the $78.50$ of distilled real images. 
At a resolution of $256\times 256$, we achieve competitive results as well. A ResNet18 can achieve performance $77.79$ in test accuracy compared to the $83.90$ of distilled real images. 

\begin{table}[tb]\setlength{\tabcolsep}{4pt}
   \centering
    \caption{Results compared to state-of-the-art. The top part shows the upper bound for the performance when the private training data is available, following results of distilled real dataset.
    The bottom part shows the results on all synthetic images or distilled synthetic images. Our performance are better than the two baselines and comparable to results of distilled real dataset in top part.}
   \resizebox{1.0\columnwidth}{!}{%
      \begin{tabular}{clccccccccc}
          &&& &\multicolumn{3}{c}{ConvNet} && \multicolumn{3}{c}{ResNet18} \\
          & \textbf{Real} & res. & \#img/class & ACC & F1 & AUC & &ACC & F1 & AUC \\
          \cmidrule{5-7}\cmidrule{9-11}
         \multirow{4}{*}{\rotatebox[origin=c]{90}{\scriptsize{upper bound}}}&\cite{yang2021medmnist,yang2023medmnist,derakhshani2022lifelonger,liu2022feature} & $64^2$ & $\sim10k$ & - & - & - & &$85.28_{\pm5.99}$ & - & $97.33_{\pm1.68}$ \\
        & Reproduced & $64^2$ & $\sim10k$ & $91.44_{\pm0.20}$ & 8$8.36_{\pm0.24}$ & $99.23_{\pm0.05}$ && $91.25_{\pm0.71}$ & $87.95_{\pm0.97}$ & $99.00_{\pm0.24}$ \\
        \cmidrule{5-7}\cmidrule{9-11}
         
         &\cite{yang2021medmnist,yang2023medmnist} & $224^2$ &$\sim10k$& - & - & - && $88.45_{\pm3.47}$ & - & $98.35_{\pm0.78}$  \\
        & Reproduced & $256^2$ & $\sim10k$ & $92.02_{\pm0.48}$ & $88.76_{\pm0.67}$ & $99.25_{\pm0.06}$ && $90.96_{\pm0.35}$ & $87.27_{\pm0.46}$ & $98.92_{\pm0.09}$  \\
         \midrule
         &\textbf{real distilled} \\
         & InfoDist & $64^2$ & 100 & $77.34_{\pm0.42}$ & $69.45_{\pm0.51}$ & $94.98_{\pm0.23}$ && $78.50_{\pm0.62}$ & $71.77_{\pm0.21}$ & $96.47_{\pm0.57}$ \\ 
         & InfoDist & $256^2$ & 100 & $81.17_{\pm0.53}$ & $73.81_{\pm0.64}$ & $96.73_{\pm0.24}$ && $83.90_{\pm0.50}$ & $77.47_{\pm0.77}$ & $97.77_{\pm0.11}$ \\
        \midrule
        \midrule
        & \textbf{Synthetic} \\
        & Reproduced & $64^2$ & 1k & $79.87_{\pm0.86}$ & $73.59_{\pm0.90}$ & $96.74_{\pm0.22}$ && $86.18_{\pm0.46}$ & $80.80_{\pm0.80}$ & $98.15_{\pm0.12}$ \\
         \midrule
         &\textbf{syn distilled} \\
         \multirow{4}{*}{\rotatebox[origin=c]{90}{\scriptsize{\textbf{results}}}}& GLaD~\cite{cazenavette2023generalizing} & $64^2$ & 1 & $40.49_{\pm1.26}$ & $28.46_{\pm1.55}$ & $74.45_{\pm0.74}$ && $46.68_{\pm1.16}$ & $34.63_{\pm0.87}$ & $75.97_{\pm0.41}$ \\
         & Random & $64^2$ &100& $61.52_{\pm0.60}$ & $54.29_{\pm0.65}$ & $89.31_{\pm0.94}$ & &$72.82_{\pm2.13}$ & $64.74_{\pm3.10}$ & $94.33_{\pm1.03}$ \\
         &InfoDist & $64^2$ & 100 & $\mathbf{69.79_{\pm1.28}}$ & $\mathbf{63.01_{\pm1.10}}$ & $\mathbf{91.15_{\pm0.35}}$ && $\mathbf{77.48_{\pm1.52}}$ & $\mathbf{70.80_{\pm1.03}}$ & $\mathbf{96.84_{\pm0.33}}$  \\
         \cmidrule{5-7}\cmidrule{9-11}
        &GLaD~\cite{cazenavette2023generalizing} & $256^2$ & 1 & $38.81_{\pm0.91}$ & $29.44_{\pm0.78}$ & $70.26_{\pm0.60}$ && $45.70_{\pm2.59}$ & $35.59_{\pm1.43}$ & $82.24_{\pm0.55}$  \\
        &Random & $256^2$ &100& $58.29_{\pm2.58}$ & $50.95_{\pm2.29}$ & $89.97_{\pm0.97}$ && $71.82_{\pm3.21}$ & $63.88_{\pm3.14}$ & $94.99_{\pm0.86}$ \\
        & InfoDist & $256^2$ & 100 & $\mathbf{65.45_{\pm1.40}}$ & $\mathbf{57.47_{\pm1.40}}$ & $\mathbf{91.45_{\pm0.43}}$ && $\mathbf{77.79_{\pm1.18}}$ & $\mathbf{72.53_{\pm1.53}}$ & $\mathbf{94.59_{\pm0.37}}$ \\
        
      \end{tabular}
    }
   \label{tab:resultcomparesota}
\end{table}

\noindent\textbf{Ablation Study.}
We provide an extensive ablation study including different embeddings (ConvNet, RestNet, UMAP) with different clustering methods in several configurations for the node selection in the Appendix. For Table~\ref{tab:resultcomparesota} we fixed the Infomap node selection metric to \emph{modular centrality}. In the Appendix we also explore the effect of using enter flow or exit flow as alternatives.
We also conducted an ablation study on hard code hyperparameters and selected the best combination to report the results.

\section{Conclusion}
In this paper, our goal is to make dataset distillation applicable in the real world because its various advantages. In our setting, privacy information is removed and data security concerns are alleviated. We initiate the process by training a latent diffusion model and generating a synthetic dataset. Subsequently, we employ cluster methods to select a smaller dataset. Classifiers are then trained on this reduced synthetic dataset, and we report the test accuracy, F1 score, and AUC score. The incorporation of a contrastive learning loss contributes to the enhancement of performance. We also select an equivalent number of real images for comparison. Our results on a distilled synthetic dataset are comparable with those on a small real dataset. 
Furthermore, our AUC score is competitive with that of the entire real dataset.

\begin{credits}
\subsubsection{\ackname} This work was supported by the State of Bavaria, the High-Tech Agenda (HTA) Bavaria and HPC resources provided by the Erlangen National High Performance Computing Center (NHR@FAU) of the Friedrich-Alexander-Universität Erlangen-Nürnberg (FAU) under the NHR project b180dc. NHR@FAU hardware is partially funded by the German Research Foundation (DFG) - 440719683. Support was also received from the ERC - project MIA-NORMAL 101083647 and DFG KA 5801/2-1, INST 90/1351-1.

\subsubsection{\discintname}
The authors have no competing interests for this work.
\end{credits}

%
% ---- Bibliography ----
%
% BibTeX users should specify bibliography style 'splncs04'.
% References will then be sorted and formatted in the correct style.
%
\bibliographystyle{splncs04}
\bibliography{Paper-0484}
\newpage
\appendix
%\section{Appendix}

\end{document}